\documentclass{article}

\usepackage{arxiv}

\usepackage[authoryear,longnamesfirst]{natbib}
\usepackage[utf8]{inputenc} 
\usepackage[T1]{fontenc}    
\usepackage{hyperref}       
\usepackage{url}            
\usepackage{booktabs}       
\usepackage{amsfonts}       
\usepackage{nicefrac}       
\usepackage{microtype}      
\usepackage{lipsum}
\usepackage{graphicx}
\usepackage{booktabs}
\usepackage{amsmath}
\usepackage{algorithm,algpseudocode}
\usepackage{boldline, makecell}
\usepackage{enumerate}
\usepackage{multirow}
\usepackage{textgreek}
\usepackage{makecell}
\usepackage{mathtools}
\newcolumntype{L}{>{\centering\arraybackslash}m{9cm}}
\usepackage{float}

\newcommand\blfootnote[1]{%
  \begingroup
  \renewcommand\thefootnote{}\footnote{#1}%
  \addtocounter{footnote}{-1}%
  \endgroup
}

\graphicspath{ {./images/} }

\DeclareUnicodeCharacter{2212}{-}

\title{An Emotion-Aware Multi-Task Approach to Fake News and Rumour Detection using Transfer Learning}

\author{
 Arjun Choudhry\footnotemark[1] \\
  Biometric Research Laboratory\\
  Delhi Technological University\\
  New Delhi, India\\
  \texttt{choudhry.arjun@gmail.com}
  \AND
 Inder Khatri\footnotemark[1] \\
  Biometric Research Laboratory\\
  Delhi Technological University\\
  New Delhi, India\\
  \texttt{inderkhatri999@gmail.com}
  \And
 Minni Jain \\
  Department of Computer Science\\
  Delhi Technological University\\
  New Delhi, India\\
  \texttt{minnijain@dtu.ac.in}  
  \And
  Dinesh Kumar Vishwakarma \\
  Biometric Research Laboratory\\
  Delhi Technological University\\
  New Delhi, India\\
  \texttt{dinesh@dtu.ac.in}
}

\begin{document}
\maketitle

\begin{abstract}
Social networking sites, blogs, and online articles are instant sources of news for internet users globally. However, in the absence of strict regulations mandating the genuineness of every text on social media, it is probable that some of these texts are fake news or rumours. Their deceptive nature and ability to propagate instantly can have an adverse effect on society. This necessitates the need for more effective detection of fake news and rumours on the web. In this work, we annotate four fake news detection and rumour detection datasets with their emotion class labels using transfer learning. We show the correlation between the legitimacy of a text with its intrinsic emotion for fake news and rumour detection, and prove that even within the same emotion class, fake and real news are often represented differently, which can be used for improved feature extraction. Based on this, we propose a multi-task framework for fake news and rumour detection, predicting both the emotion and legitimacy of the text. We train a variety of deep learning models in single-task and multi-task settings for a more comprehensive comparison. We further analyze the performance of our multi-task approach for fake news detection in cross-domain settings to verify its efficacy for better generalization across datasets, and to verify that emotions act as a domain-independent feature. Experimental results verify that our multi-task models consistently outperform their single-task counterparts in terms of accuracy, precision, recall, and F1 score, both for in-domain and cross-domain settings. We also qualitatively analyze the difference in performance in single-task and multi-task learning models.\blfootnote{*Equal Contribution}
\end{abstract}


\section{Introduction}\label{sec1}

In recent years, we have witnessed a substantial increase in the usage of social networking sites, news services, and blogs. This increase in the user base, and thus, the usage of social media has brought along an increase in the spread of \emph{fake news} and \emph{rumours}. According to the work by \citet{article}, fake news and rumours both fall under \emph{false information}. Fake news refers to the spread of false information under the pretext of being correct, while rumours are unverified pieces of information that may or may not be false. While there are noticeable differences in the semantics of a rumour and a fake news text, machine learning and deep learning approaches used for the detection of both remain largely similar. Further, both fake news and rumours have often been used interchangeably in research works. For example, \citet{ajao} proposed a sentiment-aware approach to fake news detection, but tested their approach on the PHEME \citep{pheme} dataset, predominantly used for rumour detection. Thus, both tasks can be countered with common model architectures and solutions.

Due to the deceptive nature of fake news and rumours, it is difficult for people to understand which text is real, and which is fake (or a rumour). For example, \citet{starbird} found that people on social media were unable to distinguish between rumours and non-rumours for the 2013 Boston Marathon bombings. For every tweet tweeted debunking a rumour, as many as 44 tweets were tweeted in support of it. This proves the inability of users on social media to identify fake news and rumours. Hence, we can say that social media has been unable to keep fake news and rumours off their platforms. This calls into question the credibility of social media and the web as a source of information.

Significant research has been done in recent years towards improved detection of fake news and rumours. Researchers have focused on a variety of directions to implement models that can reliably detect fake news and rumours. Some works have focused on improved classifier architectures for better learning the semantic and syntactic features that differentiate fake news texts from real news texts. For example, \citet{GNN} used Graph Neural Networks with continual learning to differentiate between the propagation patterns of fake news and real news on social media. Other works have focused on multi-modal data to extract features from the text, as well as the accompanying multimedia. This has been shown to further improve the performance of classifiers for the detection of fake news and rumours. For example, \citet{SpotFake} proposed a multi-modal framework for the detection of fake news, and found that the performance of classifiers toward the detection of fake news was significantly improved when compared with models targeting a single modality. \citet{SpotFake+} further observed improved performance by using transfer learning for large transformer models.

Some research works have also focused on the extraction of specific features from the texts, which are found to aid them in the detection of fake news and rumours. For example, \citet{kochkina} utilized veracity and stance detection features for the detection of rumours, and observed noticeably better performance from their classifiers, as compared to the classifiers trained without these features. \citet{ajao} used the sentiment ratio extracted from the text using Latent Dirichlet Allocation and concatenated the extracted vector for the sentiment ratio with the feature matrix. They observed improved performance on the PHEME 6 dataset. Some recent works have also shown the correlation between the legitimacy of a text and its emotion \citep{tweets_mit}. 

In this work, we verify the correlation between the legitimacy of a text and its intrinsic emotion and show that for the same emotion class, real news and fake news are often represented slightly differently. This work is the full paper for our abstract \citep{AAAI} and contains suitable extensions for a more thorough evaluation. Based on the correlation between fake news and emotions, we further propose a multi-class emotion-aware multi-task approach for the detection of fake news and rumours, where classifiers implicitly learn features associated with fake news or rumour detection, as well as features associated with emotions portrayed by the text, with the help of transfer learning. We annotate datasets with their emotion class labels using transfer learning. We treat fake news detection and rumour detection as similar tasks and verify our approach’s efficacy on the PHEME 9 dataset for rumour detection, FakeNews AMT dataset for multi-domain fake news detection, Celeb and Gossipcop datasets. Our experimental evaluations prove that implicit extraction of emotion-based features in a multi-task setting helps improve the model performance across a variety of model architectures and datasets, both in in-domain and cross-domain settings. Our contributions can be summarized as:
\begin{itemize}
    \item We verify the correlation between the legitimacy of a text and its portrayed emotions by dimensionality reduction. We experimentally use this correlation for improved fake news and rumour detection by proposing a multi-task framework with fake news or rumour detection as the primary task, and multi-class emotion detection as the auxiliary task.
    \item We show that for the same emotion, rumours and non-rumours are often represented slightly differently in the feature space upon dimensionality reduction. These differences will vary from dataset to dataset due to differences in the domain and other biases. We suggest that these differences can aid the model to better incorporate this correlation for improved rumours and fake news detection.
    \item We further discover that emotion-guided implicit features are more domain-independent than features extracted by models without the use of emotions, due to improved performance observed in cross-domain fake news detection.
    \item We evaluate both Ekman's and Plutchik's emotion classes for fake news and rumour detection, showing that Ekman's emotion classes lead to similar or better performance than Plutchik's emotion classes. To the best of our knowledge, no previous work on fake news detection considers Ekman's emotion classes.
    \item We evaluate the performance of the single-task and multi-task models in in-domain and cross-domain settings for different deep learning architectures and datasets. Based on the results obtained, we verify that emotion-guided implicit features extracted by our multi-task models aid in cross-domain fake news and rumour detection.
\end{itemize}

The organization of our paper is as follows: Section 2 contains the Related Works and associated literature review, Section 3 contains the Proposed Methodology, Section 4 contains the Results and Discussion, Section 5 consists of the Broader Impact and Ethical Considerations regarding our work, Section 6 conveys the Current Limitation of Our Approach, and Section 7 contains the Conclusion.  

\section{Related Work}\label{sec2}

In recent years, significant work has been done to tackle the problem of \emph{fake news} and \emph{rumour detection}. \citet{bhutani} used a combination of cosine similarity, Tf-Idf and sentiment scores as features for the detection of fake news. \citet{perez-rosas-etal-2018-automatic} presented two novel datasets, \emph{FakeNews AMT} and \emph{Celeb}, for fake news detection across multiple domains, and used hand-crafted linguistic features and an SVM model for fake news detection. \citet{saikh} treated fake news detection as a text classification task and presented two deep learning models for fake news detection on FakeNews AMT and Celeb datasets.

\citet{satya} proposed a Blockchain and LSTM-based classification technique incorporating Q-GloVe word embeddings for the detection of fake news. They efficiently used Blockchain to protect real news from adversaries to prevent its alteration. \citet{welfake} curated a new generalized dataset by combining four previously available benchmark datasets, to reduce bias associated with domains specific to each dataset. They generated embeddings from the linguistic features of the combined dataset and were used for classification using an ensemble of multiple machine learning classifiers. \citet{future_research} further presented a detailed survey of topics, challenges, and future areas of research associated with fake news detection, providing a detailed overview of dissemination, detection, and mitigation of fake news. They further focused on a specific potential future research agenda, i.e., using artificial intelligence explainable fake news credibility system. 

Some contemporary works have studied rumour detection from the direction of network sciences, focusing on its evolution and impact on users. \citet{Covid_social_networks} modelled a misinformation network for the spread of rumours during the COVID-19 pandemic. Each node in the network represented a different rumour, while the edges indicated the similarity between the rumours. This work showed that information propagation could be controlled by deleting some central nodes in the misinformation graph, and further proposed a deep learning framework to predict Twitter posts that could act as central nodes. \citet{Misinfo_mental_health}, on the other hand, studied the impact of misinformation about the COVID-19 pandemic on the mental health of people and found that individuals who consumed a higher amount of misinformation suffered from high stress and anxiety, and exhibited an increased tendency for suicidal thoughts.  

Recent research works have incorporated advanced mainstream deep learning approaches like Generative Adversarial Networks (GANs) and Transformers for fake news detection and rumour detection to train detectors in an unsupervised manner. \citet{GAN} proposed a GAN-based layered model for rumour detection, which was trained in a label-free manner without using any verified database. The Generator and the discriminator were trained adversarially in a sequential manner, where the generator produced rumours by fabricating the non-rumours, and the discriminator learnt to detect rumours and non-rumours with the help of the produced rumours. \citet{gdart} proposed a transformer-based architecture with a modified multi-head attention mechanism used to learn multiple contextual correlations for enhancing feature representation learning.

\citet{kochkina} proposed a branch LSTM-based multi-task learning model trained using veracity and stance detection as auxiliary tasks to rumour detection. The model trained with both veracity and stance as auxiliary tasks outperformed single-task models, as well as multi-task models with either veracity or stance as an auxiliary task. Building upon this work, \citet{vroc} also used multi-task learning for the rumour classification, but unlike \citet{kochkina}, they tackled all four component tasks related to rumour detection using the same model: rumour detection, rumour tracking, stance classification, and veracity classification. To improve the overall performance on the component tasks, they utilized an LSTM-based Variational-Autoencoder (VAE) to get classifier-friendly latent representations. Our proposed approach also leverages multi-task learning for rumour and fake news detection. While the above works utilize multi-task learning to tackle the different component tasks relate to rumour detection simultaneously, we exploit multi-task learning to improve implicit feature representation using emotion classification as an auxiliary task. 

Various works in fake news and rumour detection have shown the efficacy of using sentiment analysis and text polarity for fake news and rumour detection. \citet{ajao} calculated \emph{emoratio} (ratio of negative polarity to positive polarity in text) for the PHEME dataset, showing a significant difference in values for rumours and non-rumours. Augmentation of emoratio into the feature matrix showed noticeable improvements to their model's performance. \citet{ticnn} showed that real news has higher median values and lower standard deviation for both positive and negative sentiments than fake news. The median values for negative sentiment for fake news were also higher than the median values for positive sentiment. Both indicate that more negative sentiment is associated with fake news.

Some recent works have started proving that there exists an affinity between certain emotions and the legitimacy of news, and this affinity can be leveraged for better detection of fake news. \citet{Health_Fake_News} introduced a technique to develop emotion-amplified (\emph{emotionized}) text representations for the detection of fake health news articles, based on the assumption that fake and real health news articles portray a different affective character, and this can be used to aid in the detection of fake news. \citet{tweets_mit} analyzed all of the verified fake and real news stories propagated on Twitter from 2006 to 2017 and labeled them as real or fake using information extracted from six fact-checking organizations. Upon evaluating the replies on each of these stores using the lexicon curated by the National Research Council Canada (NRC) based on Plutchik's basic emotions, they detected that false stories typically garnered surprise, disgust and fear in replies, while real stories garnered anticipation, joy, trust and sadness. 

\citet{Eng_Spa} proposed an approach incorporating the modeling of users based on four characteristics, i.e. stylometry, personality, emotions, and feed embeddings, for the detection of fake news spreaders for the Profiling Fake News Spreaders on Twitter shared task at PAN 2020 \citep{PAN_2020}. They observed that stylometry, emotions, and embeddings contributed positively to the detection capability of the model. \citet{NES} proposed a common framework for novelty detection, sentiment analysis, emotion analysis, and fake news detection using a hybrid LSTM and BERT-based model. They map the emotions into two classes, \emph{emotion true} and \emph{emotion false}, which are used in their framework. \citet{Dual_1} proposed a novel Emotion-based Fake News Detection (EFN) framework utilizing the emotions extracted from both the content and the comments, i.e., the emotions portrayed by the publishers and the users respectively, apart from the content, for the detection of fake news. \citet{Dual_2} further built upon the work by \citet{Dual_1}, by extracting the \emph{Emotion Gap} between the \emph{publisher emotion} and the \emph{social emotion}, and further proposed a \emph{Dual Emotions Features} set for the detection of fake news. Most of the above emotions-based works present analytical studies about the correlation between fake news and emotions on different datasets. Only a few of them actually use emotions for detection. Further, most of these works which utilize emotions to aid in rumour or fake news detection, take the emotion labels as input or exploit the emotional inconsistencies between the main content and the comments. This makes them highly dependent on additional inputs like emotion labels, comments, and thread information for evaluation. Our work, in contrast, intends to align the feature representations according to the emotion labels in a multi-task learning setup, which does not require any additional inputs during inference time.

Some very recent works, done parallelly to our original abstract \citep{AAAI}, incorporate multi-task learning and utilize emotion recognition as an auxiliary task to aid in their main classification tasks. \citet{tweet_act} proposed a combined sentiment and emotion-aided framework for tweet act classification. They utilized the multi-modal feature extractors to obtain the textual and emoticon features. Another work \citep{temporal} investigated the impact of emotions on the temporal orientation of tweets, and further proposed a multi-task learning-based approach to leverage the influence of emotions on the temporal orientation of the text. It parallelly trained the CNN and Bi-GRU networks to learn the shared input embeddings for the primary task classifier, i.e., to detect the temporal orientation among three the classes (past, present, and future), and the auxiliary task classifier, i.e., to detect emotions among four labels (joy, sadness, anger or fear). Although these works utilize emotion classification as an auxiliary task in multi-task frameworks and are similar to our proposed approach in terms of architecture, their base problems (tweet act classification and temporal orientation of tweets) are significantly different from our task.

We observe that, contrary to the claim by \citet{tweets_mit}, rumours and non-rumours cannot always be associated with specific types of emotions, as these may vary depending on the dataset. Thus, for smaller datasets, it may not be correct to generalize rumours being associated with specific emotion classes respectively. We incorporate multi-class emotion classification to aid in fake news and rumour detection on various datasets for in-domain and cross-domain evaluation. We further use the basic emotions considered by psychologists defined by Ekman and Plutchik respectively, which form the complete set of basic emotions, in relation to fake news detection. We portray the efficacy of using multi-class emotion classification as an auxiliary task, which aids in generalizing a model for use over a wider set of datasets.

\begin{figure}[t!]
     \centering
     \includegraphics[scale=.6]{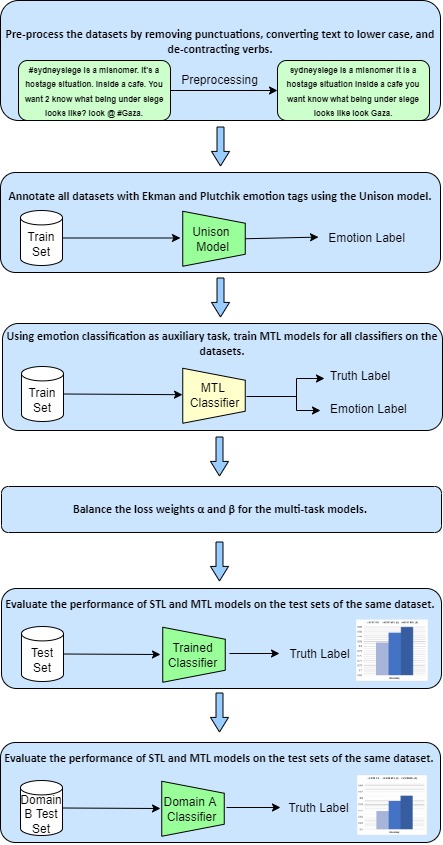}
   \caption{Flowchart representing our proposed methodology.}
   \label{Flowchart} 
\end{figure}

\section{Proposed Methodology}\label{sec3}

This section elaborates on our proposed methodology and is structured as follows: Subsection \ref{subsec1} elucidates the dataset preprocessing steps we used. Subsection \ref{subsec2} elaborates our approach for choosing the basic emotion classes for annotation and annotating the datasets. Subsection \ref{subsec3} explains the intuition behind the use of emotions as an auxiliary feature for fake news and rumour detection and shows why they aid in improving the performance of classifiers. Subsection \ref{subsec4} elaborates upon the multi-task learning framework we used for various classifiers to incorporate emotion features into the model. Subsection \ref{subsec5} elucidates the process for cross-domain analysis, the reason why we evaluated cross-domain results, and their implication. Figure \ref{Flowchart} shows a general overview of our work.

\subsection{Preprocessing the Data}\label{subsec1}

In this work, we use the PHEME 9 \citep{pheme}, FakeNews AMT \citep{perez-rosas-etal-2018-automatic}, Celeb \citep{perez-rosas-etal-2018-automatic} and Gossipcop datasets for evaluating our approach. We preprocess all the datasets using the same approach. We first convert all text to lowercase. We define a contracted verb's dictionary with general as well as specific cases that commonly occur in normal use. Using this contracted verb's dictionary, we de-contract verb forms to their root forms (eg. \textit{“I’ll”} to \textit{“I will”}. We then remove all punctuation marks and numbers. Finally, we tokenize the text and store it in a list for further use.

\subsection{Emotion Classes \& Annotating the Datasets}\label{subsec2}

Incorporating multi-class emotion classification as an auxiliary task in our multi-task framework poses two challenges: choosing a set of basic emotions, and annotating the datasets with the chosen emotion classes. There are two widely accepted theories on \textit{basic} emotion classes, proposed by \citet{plut} and \citet{ekman} respectively. \citet{plut} designed a wheel with 8 basic emotions (\textit{Joy}, \textit{Surprise}, \textit{Trust}, \textit{Anger}, \textit{Anticipation}, \textit{Sadness}, \textit{Disgust}, \textit{Fear}) based on his \emph{Ten Postulates}. Combinations of these 8 basic emotions can be used to form 56 other emotions at one intensity level. Previous psychological works \citep{tweets_mit} on the correlation between emotions and fake news have made use of Plutchik's basic emotion classes using lexicon-based approaches. During a cross-cultural study, \citet{ekman} inferred that there are 6 basic emotions (\textit{Joy}, \textit{Surprise}, \textit{Anger}, \textit{Sadness}, \textit{Disgust}, \textit{Fear}). Each emotion was considered a discrete category, rather than an individual emotion. \citet{Bann} later found Ekman's basic emotion set to be the most semantically distinct. 

We incorporate the use of Ekman's basic emotions in our work as well as Plutchik's basic emotions and see if they lead to significantly different results in the detection of fake news. Due to the unavailability of fake news datasets annotated with emotion classes, we use the \emph{Unison model}\protect\footnotemark[1] \footnotetext[1]{https://github.com/nikicc/twitter-emotion-recognition} \citep{unison} to generate both Plutchik and Ekman emotion tags to annotate our datasets. The Unison model achieved state-of-the-art results for the classification of Twitter data into their respective emotions. \citet{unison} used a dataset of 73 billion tweets to train and test their approach. Annotation using the Unison model enables us to implicitly transfer some emotion-based features from the Unison model to our classifiers when trained in a multi-task setting. While the Unison model achieves an F1-score of around 0.70 for various basic emotion classes, it is a relatively robust model trained over 73 billion tweets from a variety of domains. It is also the only available model which can annotate for both Ekman's and Plutchik's basic emotion classes, an important factor for evaluating any differences between the two emotion sets. Further, as it is trained on an extremely large real-world dataset, it is less likely to show biases than smaller models trained on smaller corpora. However, in the future, as better and more robust emotion classification models become available, we recommend their use to minimize misannotations of emotion tags and further optimize performance. We compare the performance of various classifiers in multi-task settings on both emotion sets.

\subsection{Intuition behind use of Emotions}\label{subsec3}

We incorporate emotion-guided features in our fake news detection model by implementing a multi-task framework incorporating a multi-class emotion classification task in addition to the original fake news or rumour detection task. We further hypothesize that there is a prominent correlation between the legitimacy of a text and its intrinsic Ekman's and Plutchik's emotion, and within the same emotion, real and fake texts are sometimes represented differently. We further suggest that this correlation can positively affect a classifier's capability in the detection of fake news or rumours.

\begin{figure*}[t!]
   \begin{minipage}{0.5\linewidth}
     \centering
     \includegraphics[width = 0.98\columnwidth]{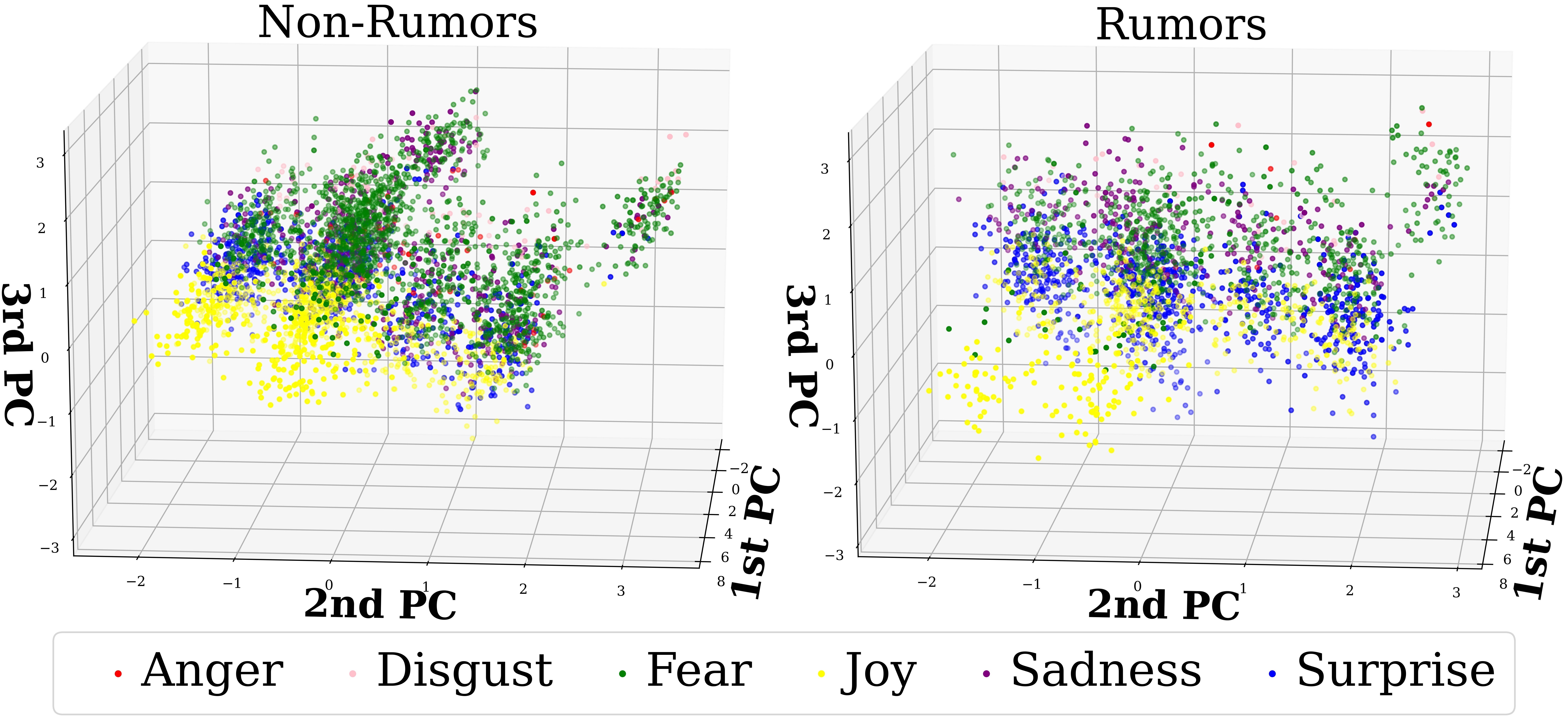}
     \\ \textit{(\textbf{A})}
   \end{minipage}\hfill
   \begin{minipage}{0.5\linewidth}
     \centering
     \includegraphics[width = 0.98\columnwidth]{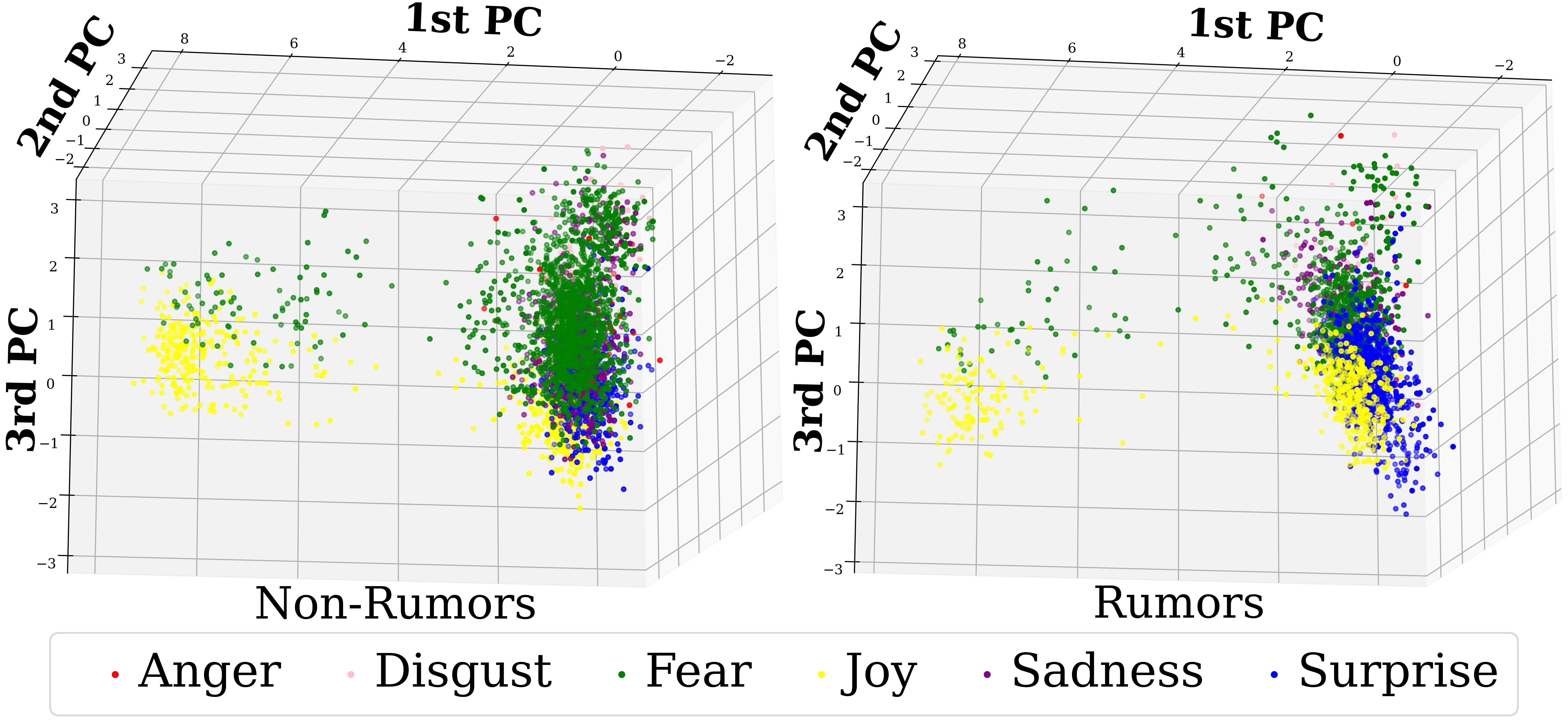}
   \\ \textit{(\textbf{B})}
   \end{minipage}
   \caption{3-dimensional Principal Component Analysis (PCA) plots for Rumours and Non-rumours from the PHEME 9 dataset, colored by Ekman emotions. \textit{(\textbf{A})}: Upper View \textit{(\textbf{B})}: Lower View}
   \label{PCAGraphs_Ekman} 
\end{figure*}

Figure \ref{PCAGraphs_Ekman} represents 3-dimensional Principal Component Analysis (PCA) graphs for the PHEME dataset, computed using embeddings generated from the Unison model for Ekman emotions. The 3-dimensional principal components are computed over the whole PHEME dataset but are plotted separately for rumours and non-rumours. If there wasn't a correlation between emotions and the legitimacy of text, then the two graphs obtained should have been nearly identical, except for the volume of sample points. However, we observe some stark differences between the plots for rumours and non-rumours. 

\begin{figure*}[t!]
   \begin{minipage}{0.5\linewidth}
     \centering
     \includegraphics[width = \columnwidth]{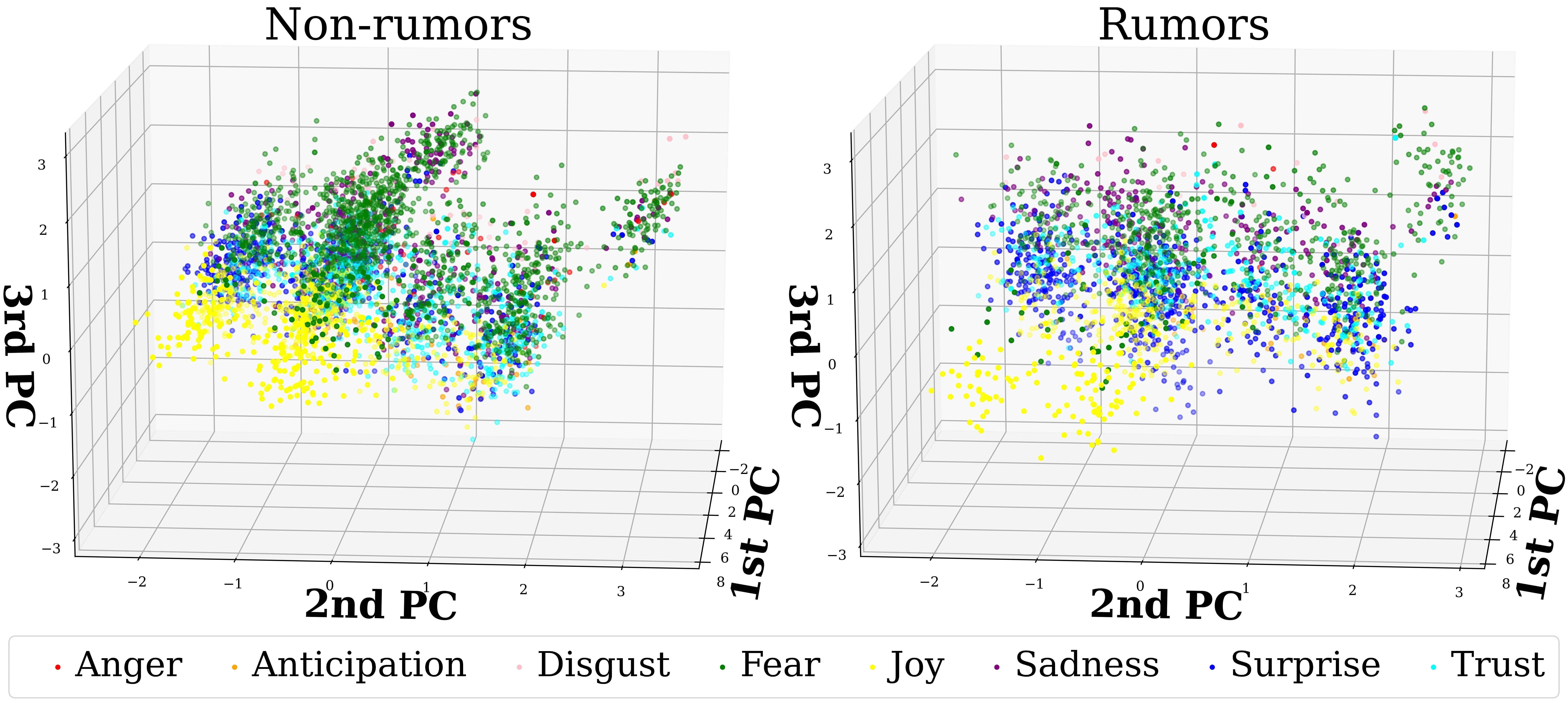}
     \\ \textit{(\textbf{A})}
   \end{minipage}\hfill
   \begin{minipage}{0.5\linewidth}
     \centering
     \includegraphics[width = \columnwidth]{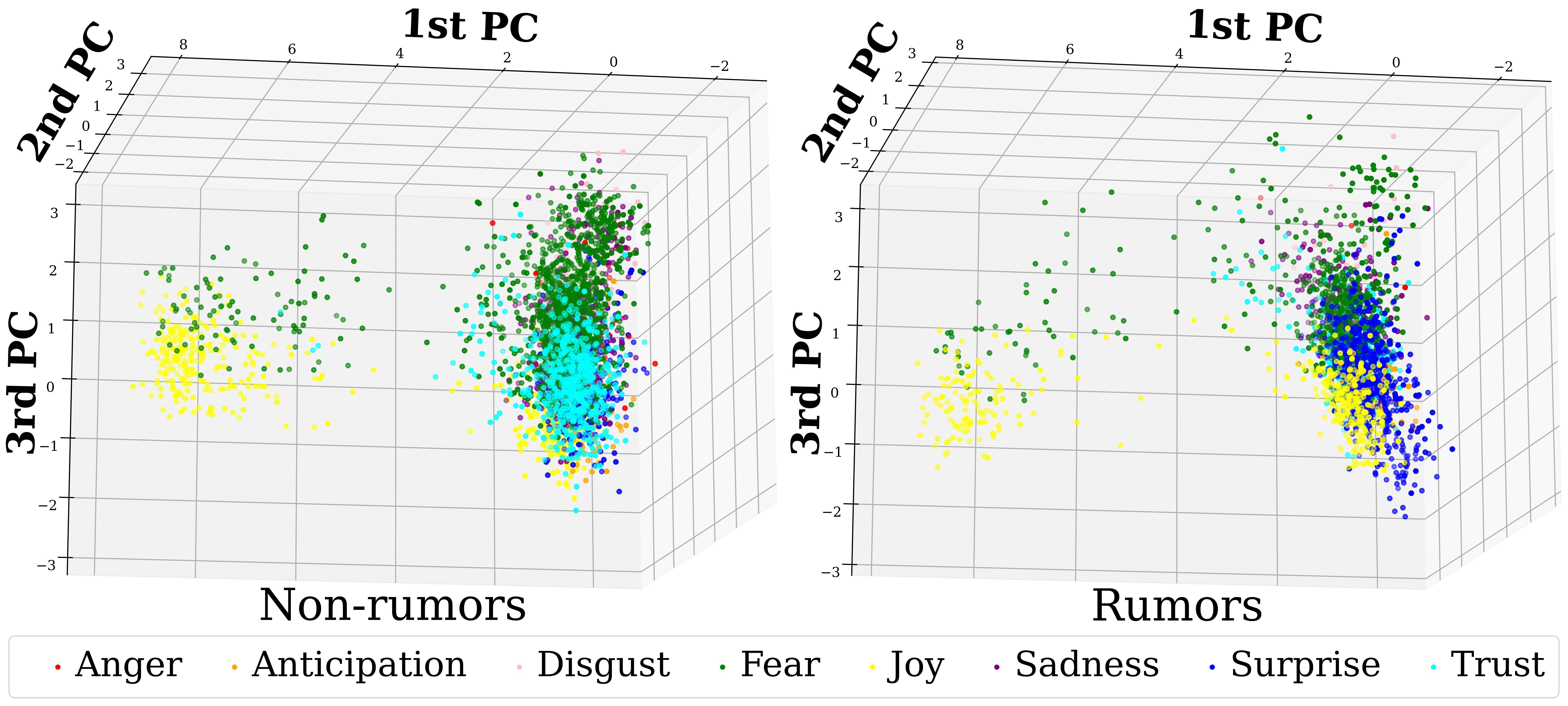}
     \\ \textit{(\textbf{B})}
   \end{minipage}
   \caption{3-dimensional Principal Component Analysis (PCA) plots for Rumours and Non-rumours from the PHEME 9 dataset, colored by Plutchik emotions. \textit{(\textbf{A})}: Upper View \textit{(\textbf{B})}: Lower View}
   \label{PCAGraphs_Plutchik} 
\end{figure*}

Non-rumours show better-formed clusters and a higher percentage of \textit{Joy} and \textit{Fear} than rumours, which are more scattered and show a higher percentage of \textit{Sadness} and \textit{Surprise}. A similar conclusion can be drawn from Figure \ref{PCAGraphs_Plutchik}, which represents 3-dimensional Principal Component Analysis (PCA) graphs for the PHEME dataset, computed using embeddings generated from the Unison model for Plutchik emotions. Non-rumours show better-formed clusters and a higher percentage of \textit{Trust} and \textit{Fear} than rumours, which are again more scattered, with a higher percentage of \textit{Sadness} and \textit{Surprise}. 

We further observe that for the same emotion, rumours and non-rumours clusters are plotted differently. Thus, even when two texts portray the same emotion, they are sometimes represented differently in the feature space for rumours and non-rumours. While the distribution of texts into emotion classes will vary from dataset to dataset, primarily due to differences in domains and various biases in each dataset, the fact that distributional and positional variations in the feature space occur between rumours and non-rumours for the same emotion should hold true for a variety of datasets, as these help form better decision boundaries for improved rumour or fake news detection in our multi-task models, leading to higher performance. The comparative non-emotion models, in comparison, perform worse and aren't able to correctly label texts as rumours or non-rumours (or fake and real news) as they are not guided by the variations in emotions between the classes. This indicates that a correlation exists between the legitimacy of a text and its underlying emotion, thereby supporting our proposition that emotion tags can aid in rumour detection and fake news detection. 

\begin{figure*}[t!]
   \begin{minipage}{\linewidth}
     \centering
     \includegraphics[width = 0.98\linewidth]{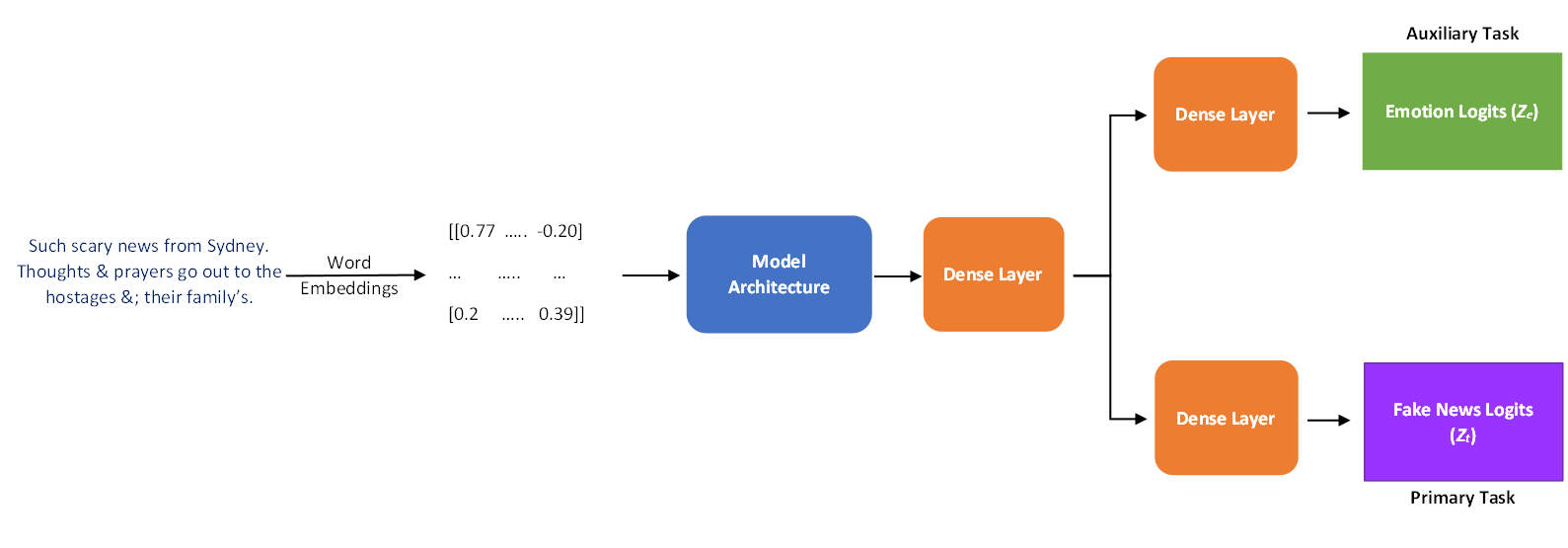}
   \end{minipage}\hfill
   \caption{Generalized multi-task learning framework with fake news detection as the primary task and multi-class emotion classification as the auxiliary task.}
   \label{MTL_Framework} 
\end{figure*}

\subsection{Multi-task Learning for Fake News Detection}\label{subsec4}

Multi-task learning (MTL) uses shared representations between the primary and auxiliary tasks for better extraction of common features, which can otherwise get ignored, and to generalise better on the primary task. We train our models to predict both the legitimacy and emotion of a text to verify the relationship between its legitimacy and emotion and evaluate them against their single-task (STL) counterparts. We further leverage the domain tags for texts in FakeNews AMT dataset and evaluate the performance of multi-task learning models predicting the legitimacy and domain of the text, to compare with multi-task learning models predicting the legitimacy and emotion of the text. Figure \ref{MTL_Framework} shows the multi-task learning architecture we use, with fake news detection as the primary task, and emotion classification as the auxiliary task.

\subsection{Cross Domain Analysis}\label{subsec5}
Previous works by \citet{perez-rosas-etal-2018-automatic} and \citet{saikh} show that classifiers exhibit a significant drop in performance when training the classifier on a dataset from one domain and testing it on a dataset from a different domain, as compared to its performance on the test set from its own domain. This is primarily due to the implicit extraction of domain-dependent features by the models, which reduces its efficacy in correctly detecting fake news from other domains. We evaluate the performance of LSTM and CNN-LSTM models in both single-task and multi-task settings for cross-domain analysis by training them on FakeNews AMT dataset and testing on Celeb dataset and vice versa. This helps us verify our hypothesis of the applicability of emotions for better generalization across domains and datasets for the detection of fake news.

\section{Experimental Analysis \& Results}\label{sec4}

This section describes the experimental analysis, results, and implications. Subsection \ref{subsec6} describes the datasets that we have used. Subsection \ref{Subsec_para_tuning} describes our $\alpha$-$\beta$ tuning approach applied to ensure optimal performance in multi-task settings. Subsection \ref{subsec7} elucidates the various standard classification models we have used with our framework to show the wide applicability of our approach. It further explains how we have balanced the loss weights of the two branches in the multi-task learning models for optimum performance. Subsection \ref{subsec8} contains the performance metrics we have used to evaluate the classifiers. Subsection \ref{subsec9} contains the results and the implications we draw from them.

\subsection{Datasets}\label{subsec6}

We have used the PHEME 9 \citep{pheme}, FakeNews AMT \citep{perez-rosas-etal-2018-automatic}, Celeb \citep{perez-rosas-etal-2018-automatic} and Gossipcop\footnotemark[2] datasets for evaluating our approach in this paper. The PHEME 9 dataset contains rumours and non-rumours posted on Twitter for 9 different events. Table \ref{table_pheme} illustrates the distribution of the PHEME 9 dataset. The FakeNews AMT dataset is a multi-domain fake news dataset covering business, education, entertainment, politics, sports, and technology, while the Celeb dataset contains news about celebrities only. The Gossipcop dataset contains entertainment news articles scrapped from the web, containing over 12,000 samples. Hereon, the terms rumour and non-rumour are used specifically in the context of the PHEME 9 dataset or its evaluation, while the term fake news is used specifically in the context of the FakeNews AMT, Celeb and Gossipcop datasets, or their evaluation.

\begin{table}[t!]
    \caption{Distribution of the PHEME 9 dataset by the event. The total number of non-rumours are 67 percent more than the number of rumours.}
    \centering
    \begin{tabular}{c|c|c|c|c}
    \hline
    \hline
        Events & Threads & Tweets & Rumours & Non-rumours \\ \hline
        Charlie Hebdo & 2,079 & 38,268 & 458 & 1,621 \\
        Sydney Siege & 1,221 & 23,996 & 522 & 699 \\
        Ferguson & 1,143 & 24,175 & 284 & 859 \\
        Ottawa shooting & 890 & 12,284 & 470 & 420 \\ 
        Germanwings-crash & 469 & 4,489 & 238 & 231 \\
        Putin missing & 238 & 835 & 126 & 112 \\
        Prince Toronto & 233 & 902 & 229 & 4 \\ 
        Gurlitt & 138 & 179 & 61 & 77 \\
        Ebola Essien & 14 & 226 & 14 & 0 \\ \hline
        Total & 6,425 & 105,354 & 2,402 & 4,023 \\
    \hline
    \hline
    \end{tabular}
    \label{table_pheme}
\end{table}

\footnotetext[2]{https://www.gossipcop.com/}

\subsection{Parameter Tuning for Multi-task Models}\label{Subsec_para_tuning}
We optimize the performance of the multi-task learning models by rebalancing the weights for the loss function for the primary and auxiliary branches of the models. We take the loss weight for the emotion classification branch as \emph{\textalpha}, and the loss weight for the fake news or rumour detection branch as \emph{\textbeta}. The relation between \emph{\textalpha} and \emph{\textbeta} is in Equation \ref{alpha-beta}:
\begin{align}
    \beta &= 1 - \alpha
    \label{alpha-beta}
\end{align}
The primary branch, tasked with the detection of the legitimacy of the text, gives a logits vector $Z_t$ of size $2 \times 1$ as output, as shown in Equation \ref{Z_t}. The auxiliary branch, which detects the Ekman or Plutchik emotion portrayed by the text, gives a logits vector $Z_e$ of size $6 \times 1$ (for Ekman emotion classes) or $8 \times 1$ (for Plutchik emotion classes) as output, as shown in Equation \ref{Z_e}.
\begin{align}\label{Z_t}
    Z_t &= \begin{bmatrix}
           Z_{t}^0 \\
           Z_{t}^1 \\
        \end{bmatrix}
\end{align}

\begin{align}\label{Z_e}
    Z_e &= \begin{bmatrix}
           Z_{e}^0 \\
           Z_{e}^1 \\
           \vdots \\
           Z_{e}^{n-1}
         \end{bmatrix}
\end{align}

$Z_e$ and $Z_t$ are further activated using the Softmax activation function to generate the probability distributions $\hat{y_t}$ and $\hat{y_e}$ for the legitimacy classes and emotion classes respectively. $\hat{y_t}$ is represented as shown in Equation \ref{y_t}
\begin{align}\label{y_t}
    \hat{y_t} &= \begin{bmatrix}
           \hat{y_t^0} \\
           \hat{y_t^1} \\
        \end{bmatrix} = Softmax(Z_t)
\end{align}
where $\hat{y_t^j}$ is defined as Equation \ref{y_t_j_hat}
\begin{align}\label{y_t_j_hat}
    \hat{y_t^j} = \frac{e^{z_t^j}} {\sum_{k=0}^1 e^{z_t^k}}
\end{align}

Similarly, for the emotion branch, we get $\hat{y_e}$ as shown in Equation \ref{y_e}
\begin{align}\label{y_e}
    \hat{y_e} &= \begin{bmatrix}
           \hat{y_e^0} \\
           \hat{y_e^1} \\
           \vdots \\
           \hat{y_e^{n-1}}
        \end{bmatrix} = Softmax(Z_e)
\end{align}
where $\hat{y_e^j}$ is defined as in Equation \ref{y_e_j_hat}.
\begin{align}\label{y_e_j_hat}
    \hat{y_e^j} = \frac{e^{z_e^j}} {\sum_{k=0}^n-1 e^{z_e^k}}
\end{align}

Further, we use cross entropy loss to penalize our model for both the fake news detection branch and the emotion detection branch, as shown in Equations \ref{loss_t} and \ref{loss_e}.
\begin{align}
    Loss_t = -\sum_{i=0}^{1} \hat{y_t^{i}}*log(\hat{y_t^{i})}
    \label{loss_t}
\end{align}

\begin{align}
    Loss_e = -\sum_{i=0}^{n-1} \hat{y_e^{i}}*log(\hat{y_e^{i})}
    \label{loss_e}
\end{align}

The total loss for the multi-task learning models is defined in Equation \ref{loss_total}.
\begin{align}
    Loss_(Total) = \alpha \times Loss_e + \beta \times Loss_t
    \label{loss_total}
\end{align}

Therefore, using Equations \ref{alpha-beta}, \ref{loss_t}, \ref{loss_e} and \ref{loss_total}, we get \ref{final_loss}.
\begin{align}
    Loss_{Total} = \alpha \times -\sum_{i=0}^{n-1} \hat{y_e^{i}}*log(\hat{y_e^{i})} \notag\\
     + (1 - \alpha) \times -\sum_{i=0}^{1} \hat{y_t^{i}}*log(\hat{y_t^{i})}
    \label{final_loss}
\end{align}

\subsection{Classification Models and Hyperparameters}\label{subsec7}
We have evaluated the performance of a variety of commonly used deep learning architectures in STL and MTL settings to show the wide applicability of our approach. Across all datasets, we have trained LSTM, CNN-LSTM, and BERT \citep{BERT} models in both MTL and STL settings. The CNN layer in the CNN-LSTM model aids in mirroring the local and position invariant features and preserving n-gram representations, while the LSTM layer helps to capture the long-term dependencies between word sequences better. For the PHEME 9 dataset, we have also evaluated the performance of CNN, CapsuleNet \citep{CapsNet} and HAN \citep{HAN} models, which have previously been used for rumour detection \citep{10.1145/3269206.3271709}. Comparison across multiple models enables us to better judge the efficacy of using emotion classification as an auxiliary task in a multi-task learning setup for fake news and rumour detection. This further ensures that our approach is applicable across a variety of architectures, and isn't just dependent on the model. We have made our code available online.\footnotemark[3]
\footnotetext[3]{https://github.com/Arjun7m/Emotion-Aware-MTL-FND-TCSS}

We used the pre-trained \emph{bert-base-uncased} model, available in the Huggingface Transformers \citep{wolf-etal-2020-transformers} library, and further fine-tuned it for our task. Additional dense layers with softmax activation have been added to the ends of each branch, to predict the legitimacy of the text and its emotion. The single-task and multi-task BERT models were trained for 5 epochs with a batch size of 32 on the PHEME 9 dataset, while for the FakeNews AMT and Celeb datasets, the BERT models have a batch size of 64, and were trained for up to 50 epochs with early stopping to prevent over-fitting.

The CNN layer in the CNN and CNN-LSTM models consists of 32 filters with a filter size of 5, followed by a max pooling layer. The LSTM layer in the LSTM and CNN-LSTM models has 100 units. The CNN, LSTM and CNN-LSTM models were trained for 5 epochs, with a batch size of 64. The HAN and CapsuleNet models were trained for 5 epochs, with a batch size of 32. All models were trained using the Adam optimizer, with a learning rate of 2e-5 and cross-entropy loss function. The models trained for evaluation on the Gossipcop dataset follow the same parameters as those for PHEME 9.

During loss weights rebalancing, we took the initial value of $\alpha$ as 0.20, and after the training and testing of a model for a given set of values for $\alpha$ and $\beta$, we incremented $\alpha$ by 0.05, consequently decreasing $\beta$ by 0.05. We trained each model for thirteen values of $\alpha$ and $\beta$, till $\alpha$ reaches 0.80, and consequently $\beta$ reaches 0.20. Figure \ref{Alpha_MTL_6} illustrates the line graphs showing the variation in the accuracy of CNN, LSTM, and CNN-LSTM models on the PHEME 9 dataset with variations in $\alpha$ and $\beta$ for Ekman's and Plutchik's emotions.

\subsection{Performance Metrics}\label{subsec8}
To compare the performance of the classifiers in single-task and multi-task settings, we calculated the metrics accuracy, precision, recall and F1-score for all models.

To calculate these metrics, we need to first determine the number of True Positives, True Negatives, False Positives, and False Negatives for our models using the confusion matrix. These metrics are defined as:

\begin{itemize}
    \item True Positive: Test samples whose labels were correctly predicted as Positive.
    
    \item True Negative: Test samples whose labels were correctly predicted as Negative.
    
    \item False Positive: Test samples whose labels were incorrectly predicted as Positive, but were in fact Negative.
    
    \item False Negative: Test samples whose labels were incorrectly predicted as Negative, but were in fact Positive.
\end{itemize}

Based on the number of True Positives (TP), True Negatives (TN), False Positives (FP), and False Negatives (FN), accuracy, precision, recall and F1-score are defined as in Equations \ref{Accuracy}, \ref{Precision}, \ref{Recall} and \ref{F1}.

\begin{align}\label{Accuracy}
    Accuracy = \frac{TP + TN}{TP + TN + FP + FN}
\end{align}

\begin{align}\label{Precision}
    Precision = \frac{TP}{TP + FP}
\end{align}

\begin{align}\label{Recall}
    Recall = \frac{TP}{TP + FN}
\end{align}

\begin{align}\label{F1}
    F1-score = \frac{2TP}{2TP + FP + FN}
\end{align}

Accuracy is used to evaluate the model performance across all classes when all classes are given equal importance. Mathematically, it represents the ratio of the number of correct predictions by a model to the total number of samples in the testing set.

Precision exhibits a model's performance in classifying a sample as positive. It is a measure of \emph{exactness} and tends to reduce if the number of samples in the minority classes is increased. Mathematically, it represents the ratio of the number of labels that are correctly predicted as positive to those that are labeled as positive.

Recall exhibits a model's performance in of correctly predicting true positives (TPs). It tends to increase if the number of samples in the minority classes is increased. Mathematically, it represents the ratio of the number of samples that are correctly predicted as positive to those that are actually positive.

F1-score acts as an alternative to accuracy, as it takes into account precision and recall scores of a model equally, but doesn't require the value of the total number of observations. Mathematically, F1-score is represented as the harmonic mean of precision and recall.

\begin{table}[b!]
    \caption{Performance evaluation on FakeNews AMT and Celeb datasets for fake news detection, using accuracy, precision, recall, and F1-Score. Multi-task learning models outperform their single-task counterparts.}
    \centering
    \begin{tabular}{c|c|c|ccccc}
    \hline
    \hline
        \textbf{Dataset} & \textbf{Model} & \textbf{Setting} & \textbf{Accuracy} & \textbf{Precision} & \textbf{Recall} & \textbf{F1}\\\hline
        
        \multirow{12}{*}{\textbf{FAMT}} & \multirow{4}{*}{\textbf{LSTM}}& STL &  0.725 & \textbf{0.829} & 0.566 & 0.673 \\
        && MTL (Ekman) &  \textbf{0.775} & 0.753 & 0.816 & \textbf{0.784} \\
        && MTL (Plutchik) &   0.758 & 0.712 &  \textbf{0.866} &  0.781 \\
        && MTL (Domain) &   \textbf{0.775} &  0.770 & 0.783 & 0.776 \\ \cline{2-7}
        &\multirow{4}{*}{\textbf{CNN-LSTM}}& STL &  0.733 & 0.725 & 0.750 & 0.737 \\
        && MTL (Ekman) &   0.758 &  0.754 &  0.766 &  0.760 \\
        && MTL (Plutchik) &   \textbf{0.766} & 0.735 &  \textbf{0.833} &  \textbf{0.781} \\
        && MTL (Domain) &   0.758 &  \textbf{0.803} & 0.683 & 0.738\\ \cline{2-7}
        &\multirow{4}{*}{\textbf{BERT}}& STL &  0.816 & 0.851 & 0.766 & 0.806 \\
        && MTL (Ekman) &  \textbf{0.875} &  \textbf{0.868} &  \textbf{0.883} &  \textbf{0.875} \\
        && MTL (Plutchik) &   0.866 &  0.866 &  0.867 & 0.866 \\
        && MTL (Domain) &   0.866 & 0.854 &  \textbf{0.883} &  0.868 \\
        \hline
        \multirow{12}{*}{\textbf{Celeb}} & \multirow{3}{*}{\textbf{LSTM}}& STL &   0.736 &  \textbf{0.780} &  0.639 &  0.702 \\
        && MTL (Ekman) &  \textbf{0.752}  &  0.758 &  \textbf{0.721} &  \textbf{0.739} \\
        && MTL (Plutchik) &  0.712 & 0.735 &  0.639 & 0.684 \\ \cline{2-7}
        &\multirow{3}{*}{\textbf{CNN-LSTM}}& STL &   0.704 &  0.740 &  0.606 &  0.666 \\
        && MTL (Ekman) &   \textbf{0.712} & 0.692 &  \textbf{0.737} &  \textbf{0.714} \\
        && MTL (Plutchik) &  0.696 &  \textbf{0.744} & 0.573 & 0.648 \\ \cline{2-7}
        &\multirow{3}{*}{\textbf{BERT}}& STL &  0.816 & 0.816 & 0.815 & 0.815 \\
        && MTL (Ekman) &   0.856 &  0.859 &  0.854 &  0.855 \\
        && MTL (Plutchik) &  \textbf{0.880} &  \textbf{0.881} &  \textbf{0.879} &  \textbf{0.879} \\
    \hline
    \hline
    \end{tabular}
    \label{results1}
\end{table}

\begin{table}[t!]
    \caption{Performance evaluation on the PHEME 9 dataset for rumour detection, using accuracy, precision, recall, and F1-score. MTL (Ekman) and MTL (Plutchik) models outperform their STL counterparts across most metrics.}
    \centering
    \begin{tabular}{c|c|c|cccc}
    \hline
    \hline
        \textbf{Dataset} & \textbf{Model} & \textbf{\makecell{\small{Setting}}} & \textbf{\makecell{\small{Accuracy}}} & \textbf{\makecell{\small{Precision}}} & \textbf{\makecell{\small{Recall}}} & \textbf{\makecell{\small{F1}}}\\\hline
        
        \multirow{18}{*}{\textbf{PHEME 9}}& \multirow{3}{*}{\textbf{CNN}}& STL &  0.870 & 0.819 &  0.826 &  0.822 \\
        & & MTL (Ekman) &  0.874 &  \textbf{0.843} & 0.800 &  0.822  \\
        & & MTL (Plutchik) &  \textbf{0.875} &  0.825 &  \textbf{0.834} &  \textbf{0.830} \\
        \cline{2-7}
        & \multirow{3}{*}{\textbf{LSTM}}& STL & 0.857 & 0.807 & 0.800 & 0.804 \\
        & & MTL (Ekman) &  0.880 &  0.835 &  \textbf{0.837} &  \textbf{0.836} \\
        & & MTL (Plutchik) &  \textbf{0.881} &  \textbf{0.847} &  0.823 &  0.835 \\
        \cline{2-7}
        & \multirow{3}{*}{\textbf{CNN-LSTM}}& STL & 0.860 & 0.786 &  \textbf{0.849} & 0.816 \\
        & & MTL (Ekman) &  \textbf{0.876} &  0.822 &  0.843 &  \textbf{0.832} \\
        & & MTL (Plutchik) &  0.872 &  \textbf{0.835} & 0.809 &  0.821 \\
        \cline{2-7}
        & \multirow{3}{*}{\textbf{CapsuleNet}}& STL & 0.853 &  \textbf{0.818} & 0.748 & 0.782 \\
        & & MTL (Ekman) &  0.858 & 0.804 &  \textbf{0.789} &  0.797 \\
        & & MTL (Plutchik) &  \textbf{0.863} &  0.817 &  0.784 &  \textbf{0.800} \\
        \cline{2-7}
        & \multirow{3}{*}{\textbf{HAN}}& STL & 0.847 & 0.790 &  0.769 & 0.779 \\
        & & MTL (Ekman) &  \textbf{0.867} &  \textbf{0.859} & 0.742 &  0.796 \\
        & & MTL (Plutchik) &  \textbf{0.867} &  0.828 &  \textbf{0.784} &  \textbf{0.805} \\
        \cline{2-7}
        & \multirow{3}{*}{\textbf{BERT}}& STL & 0.859 & 0.844 & 0.854 & 0.848 \\
        & & MTL (Ekman) &  \textbf{0.884} &  \textbf{0.873} &  \textbf{0.874} &  \textbf{0.873} \\
        & & MTL (Plutchik) &  0.874 &  0.859 &  0.870 &  0.864 \\
    \hline
    \hline
    \end{tabular}
    \label{results2}
\end{table}

\begin{table}[t!]
    \caption{Cross-domain evaluation on FakeNews AMT  and Celeb datasets using accuracy, precision, recall, and F1-score. We notice a drop in performance for all models compared to the in-domain setting. MTL models outperform their STL counterparts, indicating better generalization and domain-independent feature extraction.}
    \centering
    \begin{tabular}{c|c|c|c|cccc}
    \hline
    \hline
        \textbf{Train Set} & \textbf{Test Set} & \textbf{Model} & \textbf{Setting} & \textbf{Accuracy} & \textbf{Precision} & \textbf{Recall} & \textbf{F1}\\\hline
        
        \multirow{6}{*}{\textbf{FAMT}}& \multirow{6}{*}{\textbf{Celeb}}& \multirow{3}{*}{\textbf{LSTM}}& STL &  0.560 &  0.557 & 0.475 & 0.513 \\
        &&& MTL (Ekman) &   0.576 & 0.554 &  \textbf{0.672} &  \textbf{0.607} \\
        &&& MTL (Plutchik) &   \textbf{0.632} &  \textbf{0.641} &  0.557 &  0.596 \\ \cline{3-8}
        &&\multirow{3}{*}{\textbf{CNN-LSTM}}& STL &  0.568 & 0.559 & 0.540 & 0.550 \\
        &&& MTL (Ekman) &   \textbf{0.632} &  0.600 &  \textbf{0.737} &  \textbf{0.661} \\
        &&& MTL (Plutchik) &   \textbf{0.632} &  \textbf{0.608} &  0.688 &  0.646 \\
        \hline
        \multirow{6}{*}{\textbf{Celeb}}& \multirow{6}{*}{\textbf{FAMT}}& \multirow{3}{*}{\textbf{LSTM}}& STL &  0.558 & 0.549 & 0.650 & 0.595 \\
        &&& MTL (Ekman) &   0.591 &  0.557 &  \textbf{0.883} &  \textbf{0.683} \\
        &&& MTL (Plutchik) &   \textbf{0.608} &  \textbf{0.594} &  0.683 &  0.635 \\ \cline{3-8}
        &&\multirow{3}{*}{\textbf{CNN-LSTM}}& STL & 0.525 & 0.529 & 0.450 & 0.486 \\
        &&& MTL (Ekman) &   0.575 &  \textbf{0.563} &  0.666 &  0.610 \\
        &&& MTL (Plutchik) &   \textbf{0.583} &  0.556 &  \textbf{0.816} &  \textbf{0.662} \\
    \hline
    \hline
    \end{tabular}
    \label{results3}
\end{table}

\begin{table}[t!]
    \caption{Performance evaluation on the Gossipcop dataset for fake news detection, using accuracy, precision, recall, and F1-score. MTL (Ekman) and MTL (Plutchik) models outperform their STL counterparts across most metrics.}
    \centering
    \begin{tabular}{c|c|c|cccc}
    \hline
    \hline
        \textbf{Dataset} & \textbf{Model} & \textbf{\makecell{\small{Setting}}} & \textbf{\makecell{\small{Accuracy}}} & \textbf{\makecell{\small{Precision}}} & \textbf{\makecell{\small{Recall}}} & \textbf{\makecell{\small{F1}}}\\\hline
        
        \multirow{21}{*}{\textbf{Gossipcop}}& \multirow{3}{*}{\textbf{LSTM}}& STL & 0.806 & 0.845 & 0.932 & 0.886 \\
        & & MTL (Ekman) & \textbf{0.835} & \textbf{0.847} & 0.973 & \textbf{0.906}  \\
        & & MTL (Plutchik) & 0.833 & 0.844 & \textbf{0.975} & 0.905 \\
        \cline{2-7}
        & \multirow{3}{*}{\textbf{CNN}}& STL & 0.822 & 0.850 & 0.948 & 0.897 \\
        & & MTL (Ekman) & 0.835 & 0.850 & \textbf{0.969} & \textbf{0.906} \\
        & & MTL (Plutchik) & \textbf{0.836} & \textbf{0.853} & 0.964 & 0.905 \\
        \cline{2-7}
        & \multirow{3}{*}{\textbf{CNN-LSTM}}& STL & 0.820 & 0.842 & 0.957 & 0.896 \\
        & & MTL (Ekman) & \textbf{0.838} & \textbf{0.847} & 0.978 & \textbf{0.907} \\
        & & MTL (Plutchik) & 0.837 & 0.845 & \textbf{0.980} & \textbf{0.907} \\
        \cline{2-7}
        & \multirow{3}{*}{\textbf{CapsuleNet}}& STL & 0.824 & 0.828 & 0.977 & 0.897 \\
        & & MTL (Ekman) & \textbf{0.834} & \textbf{0.838} & 0.978 & 0.902 \\
        & & MTL (Plutchik) & 0.833 & 0.834 & \textbf{0.98}3 & \textbf{0.903} \\
        \cline{2-7}
        & \multirow{3}{*}{\textbf{HAN}}& STL & 0.864 & \textbf{0.903} & 0.927 & 0.915 \\
        & & MTL (Ekman) & 0.869 & 0.882 & \textbf{0.962} & \textbf{0.921} \\
        & & MTL (Plutchik) & \textbf{0.870} & 0.888 & 0.955 & 0.920 \\
        \cline{2-7}
        & \multirow{3}{*}{\textbf{BERT}}& STL & 0.851 & 0.803 & 0.716 & 0.745 \\
        & & MTL (Ekman) & \textbf{0.866} & \textbf{0.823} & 0.750 & 0.777 \\
        & & MTL (Plutchik) & 0.856 & 0.790 & \textbf{0.768} & \textbf{0.778} \\
        \cline{2-7}
        & \multirow{3}{*}{\textbf{RoBERTa}}& STL & 0.841 & 0.764 & \textbf{0.760} & 0.762 \\
        & & MTL (Ekman) & 0.867 & 0.833 & 0.743 & 0.774 \\
        & & MTL (Plutchik) & \textbf{0.872} & \textbf{0.836} & 0.751 & \textbf{0.782} \\
    \hline
    \hline
    \end{tabular}
    \label{results4}
\end{table}

\subsection{Results and Discussion}\label{subsec9}
We evaluated the performance of our proposed framework on three datasets and several deep learning models, using the same train-test split for each dataset across all classifiers. Table \ref{results2} illustrates the results of our experiments on the PHEME 9 dataset, Table \ref{results1} compares the results of our MTL models with their STL counterparts on the FakeNews AMT and Celeb datasets, Table \ref{results4} shows the results of our experiments on the Gossipcop dataset, and Table \ref{results3} compares the performance of our multi-task learning models with their single-task counterparts in cross-domain settings. Some interesting findings observed are:

\textbf{MTL models outperform their STL counterparts}
On comparing the performance of various classifiers in both single-task and multi-task settings, we observed an improvement in the performance of the classifiers in multi-task settings over the single-task setting across most metrics and on all datasets, verifying the correlation between the legitimacy of a text and its intrinsic emotion. The LSTM and CNN-LSTM models on the Celeb dataset unexpectedly showed slightly worse performance in the multi-task settings with Plutchik emotions compared to the single-task setting. However, the same models in the multi-task settings with Ekman's emotions outperform their single-task counterparts handily in terms of accuracy, recall, and F1 score. Further, the BERT model trained in the multi-task setting with Plutchik's emotions handily outperforms its multi-task counterpart with Ekman's emotions and its single-task counterpart across all metrics.

\textbf{MTL models with Ekman's and Plutchik's emotions perform comparably}
We observed largely comparable performance between multi-task models trained on Ekman's and Plutchik's emotions across all datasets, except for LSTM and CNN-LSTM models on the Celeb dataset, which unexpectedly performed worse with Plutchik's emotions. Multi-task models with Ekman's emotions generally performed slightly better on the FakeNews AMT, Celeb, and Gossipcop datasets while multi-task models with Plutchik's emotions performed slightly better on the PHEME 9 dataset.  However, multi-task models trained with either emotion set generally outperformed the single-task models across most metrics on all datasets. Thus, we verify the correlation between the legitimacy of a text and its portrayed Ekman emotion and concur that Ekman's emotions can also act as an alternative to Plutchik's emotions as a feature for fake news detection, particularly for fake news detection datasets.

\textbf{MTL models with domain classification as auxiliary task outperform their STL counterparts}
Classifiers trained to identify both domain and legitimacy of news (\emph{MTL (Domain)} in Table \ref{results1}) outperformed their single-task counterparts on the FakeNews AMT dataset, achieving performance comparable to those of emotion-based MTL models. This indicates a deeper correlation between the legitimacy of news and its domain. We intend to further build upon this correlation in future work for building a more generalizable framework.

\textbf{Performance drops in cross-domain settings for all models, but MTL models outperform their STL counterparts}
We observed an all-around drop in the performance of classifiers in cross-domain settings, both when trained on the FakeNews AMT dataset and tested on the Celeb dataset, and vice versa. A similar drop in performance was also observed by \citet{perez-rosas-etal-2018-automatic} and \citet{saikh}. However, our multi-task models still outperformed their single-task counterparts, implying improved extraction of shared features and better generalization. The results further indicate that intrinsic emotions aid in domain-independent implicit feature extraction for the detection of fake news.

\begin{figure*}[t!]
     \begin{minipage}{0.5\linewidth}
     \centering
     \includegraphics[width = \columnwidth]{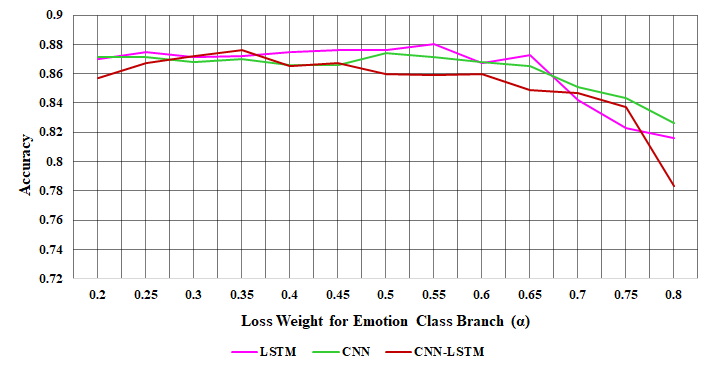}
     \\ \textit{(\textbf{A})}
   \end{minipage}\hfill
   \begin{minipage}{0.5\linewidth}
     \centering
     \includegraphics[width = \columnwidth]{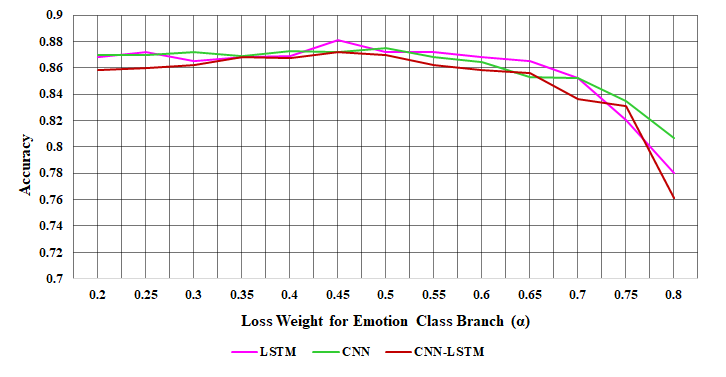}
     \\ \textit{(\textbf{B})}
   \end{minipage}\hfill
   \caption{Line graph plot for variation in accuracy of CNN, LSTM and CNN-LSTM models in multi-task setting on the PHEME 9 dataset with variation in \emph{\textalpha} (and consequently \emph{\textbeta}) for: \textit{(\textbf{A})} Ekman's emotions \textit{(\textbf{B})} Plutchik's emotions.}
   \label{Alpha_MTL_6} 
\end{figure*}

\textbf{Rebalancing weights of the loss function for fake news or rumour detection and emotion classification branches improves the performance of multi-task models}
We observed an improvement in the performance of all multi-task models on rebalancing the weights of the loss function for each branch's output. We observe from Figure \ref{Alpha_MTL_6} that the performance of the CNN, LSTM, and CNN-LSTM models on the PHEME 9 dataset reaches the maximum for \emph{\textalpha} between 0.35 and 0.55. Upon increasing the value of \emph{\textalpha} beyond that, we see a sharp decrease in the performance of the models. This can be attributed to the fact that the loss function penalizes the emotion classification branch significantly more than it penalizes the rumour detection branch (in the case of the PHEME 9 dataset). For \emph{\textalpha} = 0.80, the value of \emph{\textbeta} is 0.20. This leads to a 4:1 ratio in the weights of the loss function, in favor of the emotion classification branch, which leads to the model focusing significantly more on the performance of the emotion classification branch. 

\begin{table*}[hbt!]
    \caption{Qualitative analysis of the results obtained from single-task and multi-task learning models on the PHEME 9 dataset, where multi-task models correctly predicted the label. Label 0 corresponds to Non-rumours, while label 1 corresponds to Rumours.}
    \centering
    \resizebox{\textwidth}{!}{
    \begin{tabular}{L|c|c|c|ccc}
    \hline
    \hline
        \textbf{Text} & \textbf{Label} & \textbf{\thead{Ekman \\ Emotion}} & \textbf{\thead{Plutchik \\ Emotion}} & \textbf{STL} & \textbf{\thead{MTL \\ (Ekman)}} & \textbf{\thead{MTL \\ (Plutchik)}}\\\hline
        
        Family claims \#CorneliusGurlitt was mentally unfit. Read the news @Artlyst http://t.co/1tmcwtUYC8 http://t.co/tfH3caD1EP & 1 & Sadness & Surprise & 0 &  1 &  1\\ \hline
        Such scary news from Sydney. Thoughts \& prayers go out to the hostages \&; their family's. & 0 & Fear & Fear & 1 &  0 &  0\\ \hline
        Video footage shows police storming Sydney cafe http://t.co/3U4aWpVmdM \#SydneySiege & 0 & Surprise & Surprise & 1 &  0 &  0\\ \hline
        This is trending on French twitter \#MouradHamydInnocent started by classmates who say Mourad was in class when \#CharlieHebdo attack happened & 1 & Fear & Fear & 0 &  1 &  1\\ \hline
                
    \hline
    \hline
    \end{tabular}}
    \label{Qualitative_both_correct}
\end{table*}

\begin{table*}[hbt!]
    \caption{Qualitative analysis of the results obtained from single-task and multi-task learning models on the PHEME 9 dataset, where multi-task models with Ekman's emotions correctly predicted the label, while multi-task models with Plutchik's emotions incorrectly predicted the label. Label 0 corresponds to Non-rumours, while label 1 corresponds to Rumours.}
    \centering
    \resizebox{\textwidth}{!}{
    \begin{tabular}{L|c|c|c|ccc}
    \hline
    \hline
        \textbf{Text} & \textbf{Label} & \textbf{\thead{Ekman \\ Emotion}} & \textbf{\thead{Plutchik \\ Emotion}} & \textbf{STL} & \textbf{\thead{MTL \\ (Ekman)}} & \textbf{\thead{MTL \\ (Plutchik)}}\\\hline
        
        Putin juggling enough instability. He would make a live appearance by now to squash death/coup jitters. & 0 & Fear & Trust & 1 &  0 & 1\\ \hline
        Australian Mufti condemns hostage taking in \#Sydney - Statement \#SydneyHostageCrisis \#SydneySiege \#SydneyCafe http://t.co/n38XZVq5iu & 1 & Surprise & Trust & 0 &  1 & 0\\ \hline
        Whenever you want to justify shooting an unarmed person you always say they reached for your gun. George Zimmerman taught us that \#Ferguson & 0 & Fear & Fear & 1 &  0 & 1\\ \hline
                
    \hline
    \hline
    \end{tabular}}
    \label{Qualitative_MTL6_correct}
\end{table*}

\begin{table*}[hbt!]
    \caption{Qualitative analysis of the results obtained from single-task and multi-task learning models on the PHEME 9 dataset, where multi-task models with Plutchik's emotions correctly predicted the label, while multi-task models with Ekman's emotions incorrectly predicted the label. Label 0 corresponds to Non-rumours, while label 1 corresponds to Rumours.}
    \centering
    \resizebox{\textwidth}{!}{
    \begin{tabular}{L|c|c|c|ccc}
    \hline
    \hline
        \textbf{Text} & \textbf{Label} & \textbf{\thead{Ekman \\ Emotion}} & \textbf{\thead{Plutchik \\ Emotion}} & \textbf{STL} & \textbf{\thead{MTL \\ (Ekman)}} & \textbf{\thead{MTL \\ (Plutchik)}}\\\hline
        
        Latest footage from \#sydneysiege shows security forces throwing objects into the cafe http://t.co/HdTPHSMjUQ https://t.co/2pu3pWqHhz & 0 & Surprise & Trust & 1 & 1 &  0\\ \hline
        It just feels very odd.What'ss going on in \#Kremlin? Military coup and putin removal?I'm afraid the next Russia's leader won't be any better & 1 & Fear & Trust & 0 & 0 &  1\\ \hline
        PM Tony Abbott will hold a press conference on the \#SydneySiege in 20 minutes. & 0 & Surprise & Anticipation &  0 & 1 &  0\\ \hline
                
    \hline
    \hline
    \end{tabular}}
    \label{Qualitative_MTL8_correct}
\end{table*}

\begin{table*}[hbt!]
    \caption{Qualitative analysis of the results obtained from single-task and multi-task learning models on the PHEME 9 dataset, where multi-task models with both Ekman's and Plutchik's emotions incorrectly predicted the label. Label 0 corresponds to Non-rumours, while label 1 corresponds to Rumours.}
    \centering
    \resizebox{\textwidth}{!}{
    \begin{tabular}{L|c|c|c|ccc}
    \hline
    \hline
        \textbf{Text} & \textbf{Label} & \textbf{\thead{Ekman \\ Emotion}} & \textbf{\thead{Plutchik \\ Emotion}} & \textbf{STL} & \textbf{\thead{MTL \\ (Ekman)}} & \textbf{\thead{MTL \\ (Plutchik)}}\\\hline
        
        \#sydneysiege is a misnomer. It's a hostage situation. Inside a cafe. You want 2 know what being under siege looks like? look @ \#Gaza. & 1 & Fear & Trust &  1 & 0 & 0\\ \hline
        Gun fire exchange in Parliament Hill building \#Ottawa \#parliament \#shooting \#Canada \#BeSafeOttawa -https://t.co/9RDjNvVkJW via @martinvass2 & 1 & Joy & Trust &  1 & 0 & 0\\ \hline
                
    \hline
    \hline
    \end{tabular}}
    \label{Qualitative_both_incorrect}
\end{table*}

Upon qualitatively analyzing the results of the single-task and multi-task learning models, we found many instances where by leveraging the emotion-specific features, the multi-task learning models were able to accurately detect the legitimacy of the text. Table \ref{Qualitative_both_correct} summarises some of the tweets found in the PHEME 9 dataset that was incorrectly labelled by the majority of the single-task learning models, but using the emotion-based features, the multi-task learning models correctly predicted the labels. We observed that the multi-task learning models were able to correctly predict the legitimacy label for a sample even if the sample portrayed an emotion typically associated with the other class. This can be attributed to the fact that even within the same emotions, fake news and real news differ in their semantic and syntactic features, as shown in the PCA graphs in Figures \ref{PCAGraphs_Ekman} and \ref{PCAGraphs_Plutchik}.

We also found a few instances where multi-task learning models leveraging either emotion sets incorrectly predicted the labels, sometimes even when the single-task learning models correctly predicted the labels. These are summarised in Tables \ref{Qualitative_MTL6_correct}, \ref{Qualitative_MTL8_correct} and \ref{Qualitative_both_incorrect}. We can see from these results that multi-task learning models with Ekman's emotions can sometimes incorrectly predict rumours as non-rumours when the intrinsic emotions detected are \emph{Fear} or \emph{Joy}. Similarly, multi-task learning models with Plutchik's emotions can sometimes incorrectly predict rumours as non-rumours when the intrinsic emotions detected are \emph{Fear} or \emph{Trust}. However, the instances where single-task learning models were correct and multi-task learning models were incorrect were significantly lesser than the other way around.

\section{Broader Impact and Ethical Considerations}\label{sec5}
The spread of misinformation has become a significant problem for society with adverse consequences. Our research further improves upon previous machine learning-based fake news and rumour detection approaches by incorporating the correlated features between the legitimacy of a text and its intrinsic emotion, to more accurately detect fake news or rumours. Instead of taking the sentiment or emotion features as input to the models like in previous works, we apply a multi-task learning approach to detect fake news and rumours. This mitigates the need to generate emotion-based features, either computationally or manually, to be used as input to the fake news detection or rumour detection models during testing or deployment. Instead, the models implicitly detect the emotion-guided features and use them for the detection of fake news or rumours. Evaluation across various datasets for rumour detection as well as fake news detection, as well as on multiple deep learning architectures proves the wide applicability of our approach, both for in-domain and cross-domain settings. Our approach can further be implemented with any state-of-the-art deep learning model used for fake news detection or rumour detection, to further improve the performance of the model. Hence, systems incorporating our approach can aid in curbing the spread of false information on social media with increased accuracy. 

From an ethical point of view, our models may be affected by biases inherited from the training dataset. This may be biased toward gender, race, or a different set of people. We hope that further works lead to the creation of more robust large-scale fake news detection and rumour detection datasets annotated with emotion classes, which can help minimize biases in the data, and hence the models. However, we do not foresee our proposed approach causing any negative impact on society, directly or indirectly.

\section{The Current Limitations of Our Approach}
One limitation of our proposed approach is dependence on the Unison model for annotating the datasets with Ekman's and Plutchik's emotion tags. Despite giving state-of-the-art results for emotion classification for both Ekman's and Plutchik's emotions, the Unison model achieves a micro F1-score of around 0.70 for both Ekman's and Plutchik's emotion classes. This can sometimes lead to incorrect annotations for some texts in the datasets. These incorrect annotations can lead to contradictory information being learned by the models, leading to lower performance. While we achieved positive performance gains in our experimental evaluations, future advancements in emotion classification models as well as manually annotated fake news detection and rumour detection datasets with emotion tags can help in further improving the performance of classifiers trained to detect fake news or rumours in our proposed multi-task setting. 

\section{Conclusion}\label{sec6}
In this work, we verified the correlation between the legitimacy of a text and its intrinsic emotion. We annotated the datasets with their emotion classes using the Unison model, to help our models learn the associated emotion features by transfer learning. We used Plutchik's basic emotions, based on previous psychological studies, as well as proposed the use of Ekman's emotion classes for improved feature extraction in our classifiers. For implicit utilization of features linked to the emotions portrayed by a text, we proposed a multi-task learning approach using multi-class emotion detection as an auxiliary task for the detection of fake news and rumours on FakeNews AMT, Celeb, Gossipcop, and PHEME 9 datasets. Our experimental evaluation revealed that most multi-task models consistently outperformed their single-task counterparts in terms of accuracy, precision, recall, and F1-score for both Ekman's and Plutchik's emotions for in-domain and cross-domain settings. Our cross-domain evaluations showed that emotion-guided features are more domain generalizable, as we observed similar gains in cross-domain settings in MTL settings over STL settings as in in-domain settings. Upon qualitative analysis, we observed that MTL models showed improved performance not only for fake news samples portraying emotions normally associated with fake news and for real news portraying emotions normally associated with real news, but also for fake samples portraying emotions associated with real news and real samples portraying emotions associated with fake news. We further showed the efficacy of the use of the domain of news as another plausible feature for fake news detection on the FakeNews AMT dataset.

\bibliographystyle{cas-model2-names}

\bibliography{main}

\end{document}